\documentclass[11pt]{article}

\usepackage[final]{acl}

\usepackage{times}
\usepackage{latexsym}
\usepackage{booktabs}
\usepackage{amsmath}
\usepackage[T1]{fontenc}

\usepackage[utf8]{inputenc}

\usepackage{microtype}

\usepackage{inconsolata}

\usepackage{graphicx}

\usepackage{amssymb}

\usepackage[most]{tcolorbox}
\usepackage{xcolor}
\title{CT-FineBench: A Diagnostic Fidelity Benchmark for Fine-Grained Evaluation of CT Report Generation}

\author{
  \textbf{Ruifeng Yuan\textsuperscript{1,2,3}},
  \textbf{Wanxing Chang\textsuperscript{1,2}},
  \textbf{Weiwei Cao\textsuperscript{1,2,4}},
  \textbf{Bowen Shi\textsuperscript{1,2,5}},
\\
  \textbf{Zhongyu Wei\textsuperscript{3}},
  \textbf{Ling Zhang\textsuperscript{1}},
  \textbf{Jianpeng Zhang\textsuperscript{1,2,4}}
\\
\\
  \textsuperscript{1}DAMO Academy, Alibaba Group, China,
  \textsuperscript{2}Hupan Lab, 310023, China, \\
  \textsuperscript{3}Fudan University, China, 
  \textsuperscript{4}Zhejiang University, China,
  \textsuperscript{5}Shanghai Jiao Tong University, China\\
\\
  \small{
    \textbf{Correspondence:} \href{mailto:jianpeng.zhang0@gmail.com}{jianpeng.zhang0@gmail.com}
  }
}

\begin{document}
\maketitle
\begin{abstract}

The evaluation of generated reports remains a critical challenge in Computed Tomography (CT) report generation, due to the large volume of text, the diversity and complexity of findings, and the presence of fine‑grained, disease‑oriented attributes. Conventional evaluation metrics offer only coarse measures of lexical overlap or entity matching and fail to reflect the granular diagnostic accuracy required for clinical use. 
To address this gap, we propose \textbf{CT-FineBench}, a benchmark built from CT-RATE and Merlin to evaluate the fine-grained factual consistency of CT reports, constructed from CT-RATE and Merlin. Our benchmark is constructed through a meticulous, Question-Answering (QA) based process: first, we identify and structure key, finding-specific clinical attributes (e.g., location, size, margin). Second, we systematically transform these attributes into a QA dataset, where questions probe for specific clinical details grounded in gold-standard reports. The evaluation protocol for CT-FineBench involves using this QA dataset to query a machine-generated report and scoring the correctness of the answers. This allows for a comprehensive, interpretable, and clinically-relevant assessment, moving beyond superficial lexical overlap to pinpoint specific clinical errors. Experiments show that CT-FineBench correlates better with expert clinical assessment and is substantially more sensitive to fine-grained factual errors than prior metrics.
\end{abstract}

\section{Introduction}

The automatic generation of radiology reports from medical images, particularly Computed Tomography (CT) scans, promises to enhance the efficiency of clinical workflows. However, the clinical adoption of such systems hinges on robust evaluation. For complex and information-dense CT reports, diagnostic fidelity is paramount. This extends beyond the identification of findings to the precise characterization of their clinical attributes. A single flaw in reporting the location, morphology, or severity of a lesion can potentially compromise diagnostic accuracy. Therefore, a fine-grained evaluation metric focusing on clinical attributes is critical for CT report generation.

Existing evaluation metrics for radiology report generation can be classified into three types.
Conventional linguistic evaluation metrics like ROUGE~\cite{lin2004rouge} and BLEU~\cite{papineni2002bleu}, which are based on lexical overlap, are fundamentally inadequate for this task. Even more advanced, embedding-based metrics like BERTScore~\cite{zhang2019bertscore}, while better at capturing semantic similarity, still fail to identify and prioritize key medical information. Consequently, all these metrics often assign high scores to reports that are linguistically similar but clinically incorrect.
Recognizing this gap, recent research has moved towards more clinically-aware evaluation paradigms. One type of work focuses on entity-based metrics, such as RadGraph~\cite{jain2021radgraph} and RaTEScore~\cite{zhao2024ratescore}, which assess reports by extracting and comparing key medical entities like finding/disease and anatomical structures. However, their reliance on a limited set of coarse-grained entity and relation types means they often fail to capture the critical, fine-grained attributes that are crucial for diagnosis, particularly in complex CT reports. Another emerging approach, exemplified by metrics like GREEN~\cite{ostmeier2024green}, leverages Large Language Models (LLMs) as judges. Their reliance on general LLMs turns them into black box that offers feedback with less transparent and verifiability. Moreover, lacking inherent medical prior knowledge and a unified evaluation standard, their judgments can be unpredictable.

To overcome these limitations, we draw inspiration from the success of QA based evaluation in assessing factual consistency in general-domain~\cite{fabbri2022qafacteval}. We believe that a QA-based approach, when grounded in clinical knowledge, can provide a more objective, granular, and interpretable view for evaluating CT reports. Instead of asking a model to judge a report holistically, we can ask it specific, factual questions derived from clinical requirements.

In this paper, we introduce CT-FineBench, a new benchmark for fine-grained evaluation for CT report generation. Our core innovation is twofold: first, we shift the focus of evaluation from coarse-grained findings to their fine-grained clinical attributes (e.g., size, location, density, margin). Second, we refactor the evaluation task as a QA problem built upon these attributes, transforming the ambiguous task of report assessment into a verifiable, fact-checking process.
The benchmark is constructed through a meticulous process.
Given a report generation dataset, we first use data mining techniques to extract a set of corresponding attributes for each finding based on the reference reports. In collaboration with human annotators and medical knowledge, we then identify and structure a comprehensive set of key finding-specific attributes.
We systematically convert the report generation dataset into a large-scale QA dataset based on these structured attributes, where each question probes for a specific attribute of one findings. Finally, the evaluation protocol involves using this QA set to query a machine-generated report and measuring its quality based on the correctness of the extracted answers.

By decomposing a complex report into a checklist of fine-grained verifiable facts, CT-FineBench moves beyond both superficial text similarity, entity matching and holistic LLM judgments. It provides a comprehensive, interpretable, and clinically-relevant assessment that pinpoints specific factual errors at the attribute level. Our benchmark is built upon two well-known CT datasets, CT-RATE and Merlin, covering both chest and abdominal scans. Our experiments demonstrate that CT-FineBench aligns more closely with expert judgment on clinical details and is significantly more sensitive to fine-grained errors than existing metrics. We believe CT-FineBench will provide a more rigorous standard for model comparison and guide future research towards developing more clinically trustworthy report generation systems.

\section{Related Work}

\subsection{Radiology Report Evaluation}

The automatic evaluation of generated radiology reports has progressed along several distinct technical avenues. Initial efforts rely on lexical overlap metrics originally designed for machine translation, such as ROUGE~\cite{lin2004rouge} and BLEU~\cite{papineni2002bleu}. Follow by embedding similarity metrics like BERTScore~\cite{zhang2019bertscore}, which compute similarity based on the cosine distance between contextualized token embeddings. 
To instill clinical awareness, research has shifted towards metrics that explicitly model medical knowledge. 
CheXbert F1~\cite{smit2020chexbert} employs a medical entity extraction model for evaluation.
RadGraph~\cite{jain2021radgraph} pioneers the creation of a graph-based schema to represent entities (e.g., Anatomy, Observation) and their relations, calculating F1 scores over these structured outputs. RaTEScore~\cite{zhao2024ratescore} extends this by introducing a more structure entity typology and a synonym-aware encoding module. 
RadCliQ~\cite{yu2023evaluating} performs ensembling with multiple existing metrics for a comprehensive evaluation.
A recent paradigm employs LLM as evaluators~\cite{liu2023g,zheng2023judging}. In medical domain, GREEN~\cite{ostmeier2024green} uses an LLM trained via knowledge distillation from GPT-4 to identifies and explains clinical errors.

\subsection{CT Report Generation}

The automatic generation of CT reports is a pivotal task in medical AI. Early deep learning approaches for radiology report generation adapted the encoder-decoder framework from image captioning~\cite{navab2015medical,harzig2019addressing}, employing a CNN as encoder and a LSTM as decoder. More recently, Transformer-based architectures have become predominant~\cite{moor2023med,li2023llava}. With the development of CT report datasets~\cite{hamamci2024developing,blankemeier2024merlin}, models focusing on CT have emerged. 
CT2Rep~\cite{hamamci2024ct2rep} proposes to directly use a 3D vision encoder to generate CT reports.
CT-CHAT~\cite{hamamci2024developing} adapted the LLaVA framework, both demonstrating the effectiveness of large-scale language models in 3D CT understanding.
Med3DVLM~\cite{xin2025med3dvlm} presents a efficient 3D vision-language model that better aligns image features with
text embeddings.
Despite focusing on chest CT, Merlin~\cite{blankemeier2024merlin} tends to investigate CT report generation on abdominal CT.
Beyond these specialized models, broader multi-task medical models have also emerged, designed to address a wide array of medical tasks that include CT report generation~\cite{wu2025towards,xu2025lingshu,jiang2025hulu}.

\section{Method}

\begin{figure*}[ht]
    \centering
    \includegraphics[width=1\linewidth]{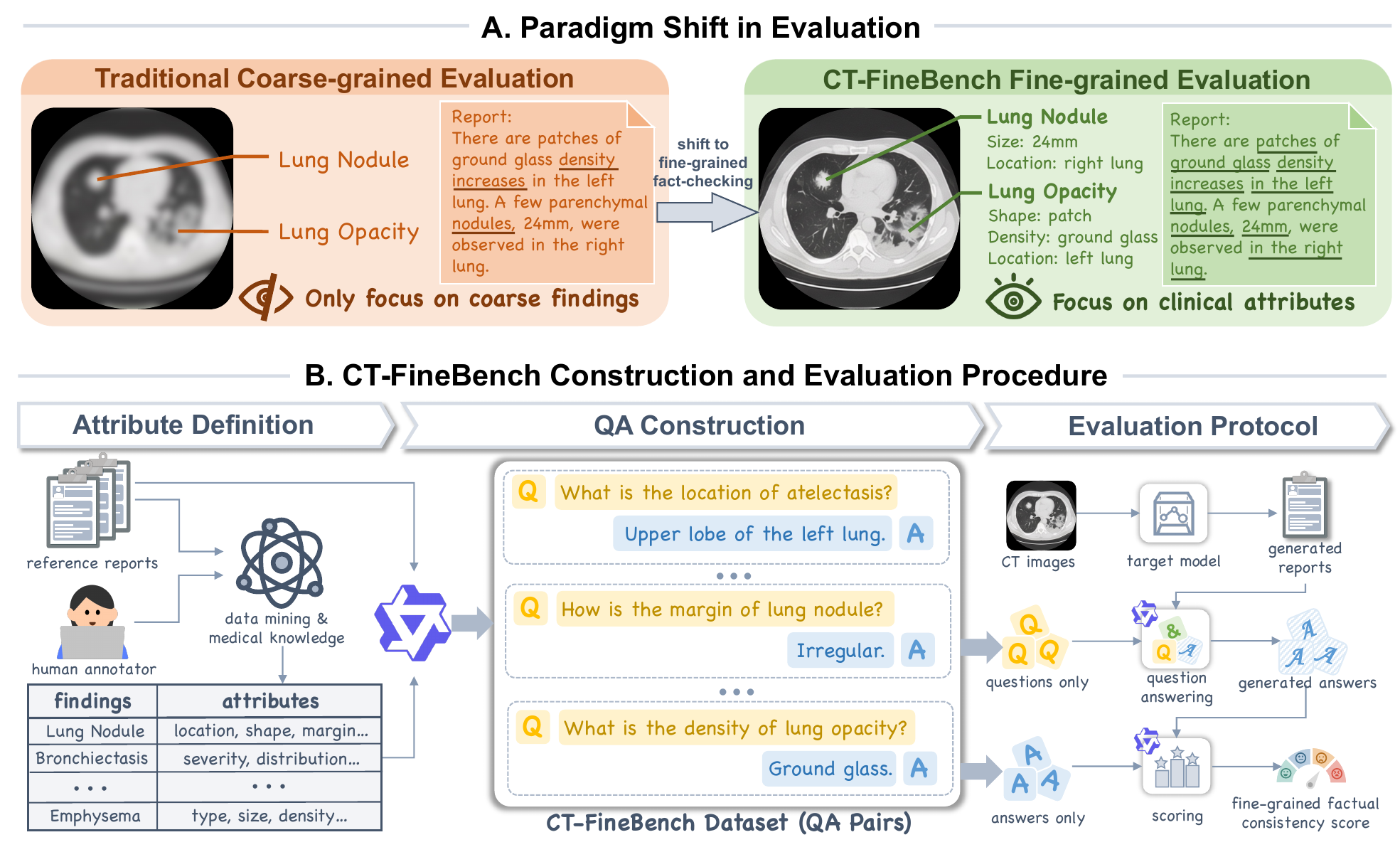}
    \caption{The framework of CT-FineBench.}
    \label{fig:Framework}
\end{figure*}

\subsection{General Pipeline}

Given a reference CT report, denoted as $x$, and a candidate report generated by a model, denoted as $\hat{x}$, our objective is to define a new metric, $\text{Score}(x, \hat{x})$, that evaluates the fine-grained factual consistency of $\hat{x}$ by verifying its key clinical attributes.

As shown in Figure~\ref{fig:Framework}, our pipeline comprises three primary phases: an offline Attribute Definition phase and QA Construction phase, followed by an online Evaluation phase.

First, in the offline phases, we construct our core Question Answering benchmark, $D_{\text{QA}}$. This is a knowledge-driven process, denoted as $\Phi_{\text{Build}}$. We first analyzes a corpus of reference reports to identify clinically significant fine-grained attributes for each critical findings (e.g., location, size, density). These structured attributes are then systematically transformed into a dataset of question-answer pairs:
\begin{equation}
    D_{\text{QA}} = \Phi_{\text{Build}}(\{x\}) = \{ (q_i, a_i) \}
    \label{eq:build}
\end{equation}
where $q_i$ is a question that probes a specific attribute (e.g., "What is the size of the lesion?") and $a_i$ is the ground-truth answer extracted from the corresponding reference report.

Second, in the online evaluation phase, we use the pre-constructed benchmark $D_{\text{QA}}$ to assess a given candidate report $\hat{x}$. This evaluation process, $\Phi_{\text{Eval}}$, is formulated as follows:
\begin{equation}
    \text{Score}(x, \hat{x}) = \Phi_{\text{Eval}}(\hat{x}, D_{\text{QA}}(x))
    \label{eq:eval}
\end{equation}
where $D_{\text{QA}}(x)$ is the subset of question-answer pairs corresponding to the reference $x$. The $\Phi_{\text{Eval}}$ module itself contains two major components: a Question Answering module and an Answer Comparison module. For each question $q_i$ in $D_{\text{QA}}(x)$, the QA module extracts a predicted answer $\hat{a}_i$ from the candidate text $\hat{x}$. This prediction is then evaluated against the ground-truth answer $a_i$ by the comparison module. The final score is the aggregated result of these comparisons, reflecting the accuracy of the candidate report at the attribute level.

\subsection{Benchmark Construction}

The construction of \texttt{CT-FineBench} is a meticulous, knowledge-driven process designed to transform unstructured clinical text into a structured, verifiable QA benchmark. This offline phase, denoted as $\Phi_{\text{Build}}$ in Section 3.1, consists of two primary stages: Attribute Definition and Question-Answer Pair Construction.

\paragraph{Attribute Definition.}
The foundation of our benchmark is a fine-grained, structured schema of clinical attributes for key findings in CT reports. We develop this schema through a multi-step, human-in-the-loop process:

We first apply data mining techniques, Named Entity Recognition (NER), on the reference reports from a target datasets. Given a report with a set of positive findings, we use LLM to extract multiple triplets $(finding, attribute, content)$ for each positive finding. This step automatically extracts a large vocabulary of findings (e.g., lung nodule, atelectasis) and their associated descriptive terms.

To refine the noisy and redundant raw extracted triplets, we first group them by their finding and attribute components, and discard any group with a frequency below a predefined threshold. Then they are reviewed and structured by human annotators. For each finding, we aim to establish a set of clinical critical attributes, ensuring there is minimal overlap and ambiguity between them. With the help of medical knowledge, the annotators are required to follow the four steps to process the attribute group of one finding: (1) Remove: remove the attributes that are clinically irrelevant. (2) Split: split general attributes like "feature" into more specific ones. (3) Merge: merge the attributes that are synonyms. (4) Comment: the annotators add an explanation for each attribute and collect a set of examples for it. This process establishes a hierarchical schema, where each finding is associated with a set of clinically relevant attributes. For example, a "lung nodule" finding link to attributes such as Location, Shape, Density, and Margin.

\paragraph{Question-Answer Pair Construction.}
With the structured attribute schema in place, we systematically convert the entire report generation dataset into a large-scale QA dataset.


For an input reference report, we generate a set of QA pairs based on finding-attribute pairs in our schema. It is worth noticing that we only use the corresponding finding-attribute pairs of the positive findings of a report. This generation process is automated using a powerful LLM guided by few-shot prompting. If a report does not mention a specific attribute, the related QA pair will be removed. 
To complement the fine-grained, attribute-level QA pairs, we introduced a set of QA pairs concerning the existence of the findings. This ensures our evaluation benchmark assesses factual consistency at both coarse-grained and fine-grained granularities.

The final output of this phase is our benchmark $D_{\text{QA}}$, a large collection of (question, ground-truth answer) pairs, each grounded in a specific clinical fact from a reference report.

\subsection{Evaluation Procedure}

The online evaluation phase, $\Phi_{\text{Eval}}$, leverages the constructed benchmark $D_{\text{QA}}$ to score a candidate report $\hat{x}$. The procedure is designed to be fully automated. It involves two main steps: Answer Extraction and Answer Comparison.

\paragraph{Answer Extraction.}
For each candidate report $\hat{x}$, we retrieve the corresponding set of questions $\{q_i\}$ from $D_{\text{QA}}(x)$. We then employ a Question Answering module, $\Phi_{\text{QA}}$, to process the candidate report. For each question $q_i$, the module is tasked with finding and extracting the most plausible answer span from the text of $\hat{x}$.
\begin{equation}
    \hat{a}_i = \Phi_{\text{QA}}(q_i, \hat{x})
\end{equation}
If the QA module determines that the question cannot be answered from the provided text (i.e., the attribute is not mentioned in the candidate report), it outputs a special token, $\hat{a}_i = \texttt{[NULL]}$.

\paragraph{Answer Comparison.}
The core of the evaluation lies in comparing the extracted answer $\hat{a}_i$ with the ground-truth answer $a_i$. This comparison, performed by the $\Phi_{\text{Compare}}$ module, assigning a score of 0, 0.5, or 1. This comparison is made type-aware by leveraging a pre-defined prompt:
\begin{itemize}
\item \textbf{Location/Categorical Attributes:} For attributes with a defined set of values (e.g., density: solid'', ground-glass''), we use a graded scoring method. A full score (1.0) is awarded for a synonym-aware exact match. A partial score (0.5) is assigned for answers that are partially correct or over-specified. Completely incorrect answers receive a score of 0.
\item \textbf{Numeric Attributes:} For quantitative attributes (e.g., size, density), the comparison function first parses and standardizes both numeric values and units. A score is then assigned based on the relative error: a full score (1.0) for an error below 10\%, a partial score (0.5) for an error between 10\% and 30\%, and a score of 0 for errors of 30\% or greater. The numeric attributes are a heuristic design, but can be easily adjusted by changing the related prompt. 
\item \textbf{Absence of Finding:} If the model predicts \texttt{[NULL]}, this is considered an error of omission (false negative) and receives a score of 0.
\end{itemize}

\paragraph{Final Score.}
The final CT-FineBench score is calculated as the average score over all question-answer pairs for a given report.

\subsection{Implementation Details}
In this section, we introduce the implementation details for benchmark construction and evaluation procedure. First, we use the whole target dataset including train set and test set for attribute definition to obtain a more comprehensive view of attribute schema. The filter threshold in attribute definition is set to 50. For the LLM used for NER in attribute definition and QA pair construction, we adopt Qwen3-Max~\cite{yang2025qwen3} with carefully designed prompt. In terms of the human annotation in attribute definition, we employ 4 annotators and each of them is individually responsible for annotating a portion of the data. These human annotators can leverage LLMs, such as Gemini-2.5-Pro~\cite{comanici2025gemini} and GPT-5\cite{openai2025gpt5}, to acquire the necessary clinical knowledge for annotation. We employ Qwen3-Max for $\Phi_{\text{QA}}$ and $\Phi_{\text{Compare}}$, while also evaluating the capability of smaller models, such as Qwen3-7b, for these same tasks in the experiment.

\section{CT-FineBench}
\label{sec:ct_factbench}

Following the methodology described in Section 3, we construct CT-FineBench, a comprehensive QA benchmark for evaluating the fine-grained accuracy of generated CT reports.

\subsection{Source Datasets}
To ensure broad applicability across different anatomical regions and clinical scenarios, CT-FineBench is built upon two distinct, publicly available CT image report paired datasets.

CT-RATE~\cite{hamamci2024developing} is a large-scale dataset focusing on chest CT scans. It contains 24109/1564 (train/test) image-report pairs, covering 18 findings. It also contains the positive finding labels for all the image-report pairs. 
Merlin~\cite{blankemeier2024merlin} is incorporated to broaden the scope of CT-FineBench to abdominal pathologies. This dataset, focusing on abdominal CT scans, provides 15175/5018/5082 (train/val/test) reports and 30 findings annotated with positive labels.

\subsection{Benchmark Statistics}

The final CT-FineBench is a collection of question-answer pairs derived from the reference reports of the source test datasets. Table~\ref{benchmark_stats} provides a comprehensive statistical overview. In total, our benchmark comprises over 44268 QA pairs derived from 6646 reports, covering a diverse set of clinical findings and their attributes. Moreover, based on the average QA pairs per report, we can observe that CT-RATE provides more detailed CT reports than Merlin.

Beyond CT-FineBench, which is designed for evaluation, we also construct a large-scale set of fine-grained QA pairs on the training splits of the source datasets, which we term CT-FineData. CT-FineData contains over 439665 QA pairs derived from 44302 reports. This parallel training corpus is a crucial component of our contribution and can be used to improve the report generation models in future work. 

\begin{table}[ht]
\centering
\resizebox{\columnwidth}{!}{%
\begin{tabular}{@{}lccc@{}}
\toprule
\textbf{Statistic} & \textbf{CT-RATE} & \textbf{Merlin}  \\ \midrule
Number of Reports & 1564 & 5082  \\
Number of Unique Attributes & 94 & 89  \\
Avg. Attributes per Finding & 5.2 & 3.0  \\
Total QA Pairs & 24148 & 20120 \\
Avg. QA Pairs per Report & 15.4 & 4.0  \\
\bottomrule
\end{tabular}%
}
\caption{Key statistics of the CT-FineBench dataset, broken down by its source datasets.}
\label{benchmark_stats}
\end{table}

\begin{figure}[ht]
    \centering
    \includegraphics[width=1\columnwidth]{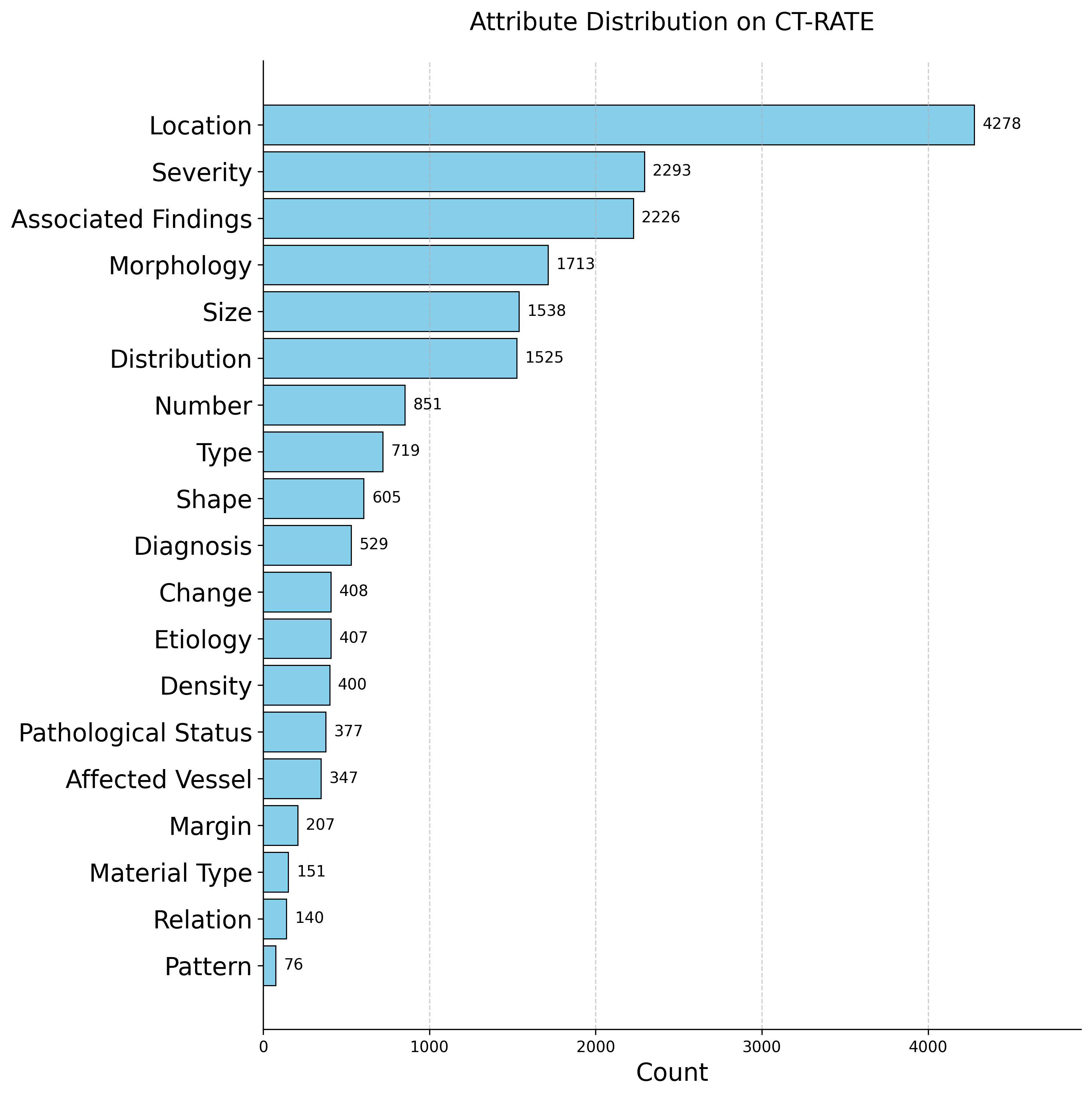}
    \caption{Distribution of QA pairs by attribute type.}
    \label{fig:attribute_distribution}
\end{figure}

To illustrate the clinical and granular focus of our benchmark, we analyze the distribution of its content.  Figure~\ref{fig:attribute_distribution} visualizes the core contribution of our work: the focus on fine-grained attributes. The chart shows the distribution of QA pairs categorized by the type of attribute they probe. A significant portion of questions relate to location, size, and other descriptive attributes, demonstrating the benchmark's ability to perform detailed, multi-faceted evaluation beyond simple entity presence.

\begin{table*}[ht]
\centering
\label{tab:qa_examples}
\resizebox{\textwidth}{!}{%
\begin{tabular}{@{}llll@{}}
\toprule
\textbf{Finding} & \textbf{Attribute} & \textbf{Question} & \textbf{Ground-Truth Answer} \\ \midrule
Lung Opacity & Location & Where is the lung opacity in the report? & Left lung \\
Lung Nodule & Size & What is the diameter of the lung nodule in millimeters? & 12 mm \\
Atelectasis & Type & What type of atelectasis is observed? & Compressive \\
Hiatal Hernia & Type & What is the classification of the hiatal hernia based on this report? & Sliding type \\
Emphysema & Density & What is the density characteristic of the emphysema in this CT report? & Diffusely clear ground glass \\
Consolidation & Margin & What is the characteristic of the margin of the consolidation in the right lung? & Irregular \\
\bottomrule
\end{tabular}%
}
\caption{Examples of fine-grained QA pairs in CT-FineBench.}
\label{qa_example}
\end{table*}

To make our methodology more concrete, Table~\ref{qa_example} presents several examples of the final QA pairs generated for \texttt{CT-FineBench}.

\section{Experiments}

\subsection{Baseline Metrics and Models}
We adopt the following metrics as the comparison of our CT-FineBench.
BLEU-2~\cite{papineni2002bleu} measures the precision of generated text by comparing 2-gram overlap.
ROUGE-L~\cite{lin2004rouge} measures the longest common sequence of words between a candidate and a reference report.
BERTScore~\cite{zhang2019bertscore} utilizes a pretrained BERT model to calculate the similarity of word embeddings between candidate and reference texts.
RadGraph F1~\cite{jain2021radgraph} extracts the radiology entities and relations for Chest Xray modality and computes the F1 score on the entity level.
RaTEScore~\cite{zhao2024ratescore} comparing clinically important medical entities on findings level, handling synonyms and negations via entity embeddings.
GREEN~\cite{ostmeier2024green} use LLM to evaluate the medical report by identifying and explaining clinically significant errors.

We evaluate the outputs of multiple models on CT report generation using CT-FineBench, assessing performance on both the CT-RATE and Merlin dataset. The models include RadFM~\cite{wu2025towards}, CT-Chat~\cite{hamamci2024developing}, Merlin~\cite{blankemeier2024merlin}, Hulu-Med~\cite{jiang2025hulu}.

\subsection{Main Results on Baseline Models}

\begin{table*}[ht]
    \centering
    \resizebox{\textwidth}{!}{%
        \begin{tabular}{lcccccc|c}
            \toprule
            \textbf{CT-RATE} & \textbf{BLEU-2} & \textbf{ROUGE-L} & \textbf{BERTScore} & \textbf{RadGraph} & \textbf{RaTEScore} & \textbf{GREEN} & \textbf{CT-FineBench} \\
            \cmidrule(r){1-8}
            RadFM     & 4.1 & 12.0 & 80.6 & 2.3 & 40.7 & 3.2 & 4.4 \\
            Hulu-Med  & 11.5 & 20.0 & 84.2 & 9.5 & 49.8 & 15.3 & 12.2 \\
            CT-CHAT    & 29.0 & 35.6 & 87.5 & 21.5 & 65.2 & 35.8 & 15.8 \\
            \cmidrule(r){1-8}
            CT-RATE-pos & 28.0 & 34.3 & 89.5 & 22.2 & 77.3 & 56.1 & 74.5 \\
            CT-RATE-neg & 70.0 & 75.3 & 95.0 & 42.9 & 76.2 & 19.2 & 39.1 \\
            \midrule
            \textbf{Merlin} & \textbf{BLEU-2} & \textbf{ROUGE-L} & \textbf{BERTScore} & \textbf{RadGraph} & \textbf{RaTEScore} & \textbf{GREEN} & \textbf{CT-FineBench} \\
            \cmidrule(r){1-8}
            RadFM     & 3.0 & 9.7 & 80.0 & 2.2 & 36.1 & 1.3 & 1.0 \\
            Hulu-Med  & 2.0 & 12.0 & 81.8 & 4.5 & 41.0 & 9.1 & 4.8 \\
            Merlin    & 9.0 & 27.4 & 85.0 & 18.7 & 64.9 & 30.2 & 22.4 \\
            \cmidrule(r){1-8}
            Merlin-pos & 29.8 & 42.3 & 88.2 & 53.3 & 81.2 & 17.6 & 86.9 \\
            Merlin-neg & 60.5 & 70.1 & 93.3 & 63.4 & 80.0 & 1.3 & 45.6 \\
            \bottomrule
        \end{tabular}%
    }
    \caption{The baseline result and sensitivity analysis experiment.}
    \label{main_result}
\end{table*}

We first evaluate a suite of baseline report generation models using our proposed CT-FineBench on both the CT-RATE and Merlin test sets in Table~\ref{main_result}.

In addition, considering a key limitation of existing metrics is their insensitivity to small but clinically critical errors. We design a sensitivity analysis experiment to evaluate CT-FineBench's ability to overcome this flaw in Table~\ref{main_result}. Our analysis is twofold, targeting both factual divergence and lexical variance:

\paragraph{Adversarial Report:} We create a set of adversarial examples (CT-RATE-neg and Merlin-neg) by introducing fine-grained clinical errors into reference reports, which serve as ``near-perfect'' but factually incorrect generated reports. Here, we only focus on fine-grained attribute errors such as location or size, and we ignore negation errors on coarse-grained findings for a more focused analysis. This process is achieved by using a prompted LLM to make minimal textual changes for maximum clinical impact. An ideal metric should assign a relatively low but non-zero score to reflect the preserved correct findings.
 
\paragraph{Paraphrased Report:} We also construct a set of reports that are factually consistent but lexically diverse (CT-RATE-pos and Merlin-pos). Similarly, we prompt the LLM to rewrite the reference reports using different phrasing, sentence structures, and synonyms, while strictly preserving all clinical facts. A robust metric should assign them a score that is close to 1.

The main results are presented in the top half of Table~\ref{main_result}. As shown, CT-FineBench reveals significant performance differences among the models, providing a clearer and more granular assessment than what can be inferred from traditional lexical metrics. An overall observation is that the absolute scores for all models on CT-FineBench are relatively low. This indicates that even state-of-the-art report generation methods struggle with fine-grained factual accuracy. This underscores a critical gap in current generation capabilities and suggests that the path towards producing fully trustworthy, clinically-reliable reports is still long.

The results of sensitivity analysis are shown in the bottom half of Table~\ref{main_result}. An ideal metric should maintain a high score on paraphrased reports (-pos), while drops significantly on adversarial reports (-neg). However, for lexical metrics like BLEU-2 and Rouge-L show an opposite trend. This suggests they are completely confused by the lexical changes on the reports and ignore the critical clinical consistency. Metrics including BERTScore, RadGraph, and RaTEScore achieve similar results for the two types of reports, indicating that they are not sensitive to fine-grained clinical errors. As a pure LLM-based approach, GREEN exhibits performance instability, performing relatively well on CT-RATE but struggling on Merlin. CT-FineBench is the only metric that behaves as desired. This demonstrates its dual ability to be robust to lexical variance while remaining highly sensitive to critical errors.


\subsection{Correlation Result}

\begin{table*}[ht]
    \centering
        \begin{tabular}{l ccc ccc}
            \toprule
            \textbf{Metric} & \multicolumn{3}{c}{\textbf{CT-RATE}} & \multicolumn{3}{c}{\textbf{Merlin}} \\
            \cmidrule(lr){2-4} \cmidrule(lr){5-7}
             & \textbf{Pearson $\tau$} & \textbf{Kendall $\tau$} & \textbf{Spearman $\tau$} & \textbf{Pearson $\tau$} & \textbf{Kendall $\tau$} & \textbf{Spearman $\tau$} \\
            \midrule
            BLEU-2       & 0.495 & 0.353 & 0.479 & 0.168 & 0.089 & 0.114 \\
            ROUGE-L      & 0.170 & 0.115 & 0.158 & 0.124 & 0.070 & 0.097 \\
            BERTScore    & 0.349 & 0.262 & 0.342 & 0.311 & 0.217 & 0.293 \\
            RadGraph     & 0.163 & 0.137 & 0.171 & \textbf{0.369} & 0.251 & 0.341 \\
            RaTEScore    & 0.521 & 0.320 & 0.434 & 0.334 & 0.227 & 0.303 \\
            GREEN        & 0.111 & 0.063 & 0.088 & 0.309 & 0.179 & 0.228 \\
            CT-FineBench & \textbf{0.622} & \textbf{0.378} & \textbf{0.490} & 0.326 & \textbf{0.306} & \textbf{0.401} \\
            \bottomrule
        \end{tabular}%
    \caption{Correlation results of evaluation metrics with human result on CT-RATE and Merlin Dataset.}
    \label{ctrate_merlin_correlation_descriptive}
\end{table*}

To validate that CT-FineBench aligns with human judgment, we correlated its outputs with assessments from human experts in Table~\ref{ctrate_merlin_correlation_descriptive}. A high correlation would provide strong evidence that our benchmark is a reliable proxy for human assessment. We randomly sample 100 generated reports from CT-RATE (generated by CT-CHAT) and Merlin (generated by Merlin) respectively. Two experts are commissioned to independently evaluate the factual accuracy of each report on a 10-point scale relative to the ground-truth report. The final human score is the average of the two annotators' ratings. We then compute the Pearson/Kendall/Spearman correlation coefficient between the average human scores and the scores produced by each automatic metric, including our CT-FineBench.

As shown in Table~\ref{ctrate_merlin_correlation_descriptive}, CT-FineBench achieves the highest correlation with human judgments across nearly all measures on both datasets. Specifically, on the CT-RATE dataset, it achieves a Pearson's $\tau$ of 0.622, substantially outperforming the next-best metric, RaTEScore (0.521). This suggests that our fine-grained, attribute-based QA approach more closely mirrors the cognitive process of an expert verifying a checklist of critical facts, compared to methods based on entity matching or holistic text similarity. While its lead on the Merlin dataset is more modest, its consistent performance across both chest and abdominal domains underscores its robustness and validity.

\subsection{Inter-Metric Correlation}

\begin{figure}[ht]
    \centering
    \includegraphics[width=1\columnwidth]{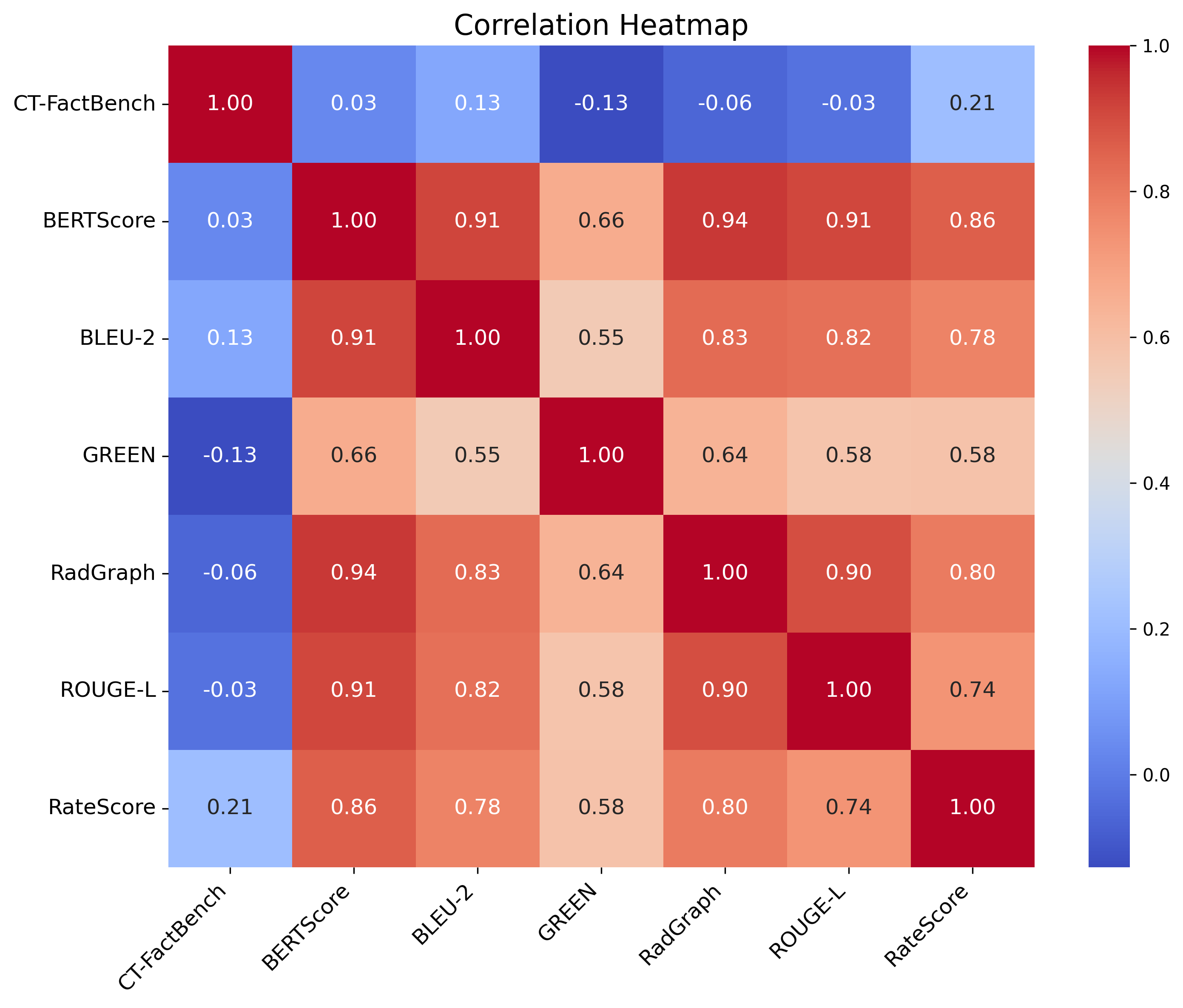}
    \caption{The heatmap of inter-metric correlation.}
    \label{metric_correlation}
\end{figure}

In Figure~\ref{metric_correlation}, we analyze the Pearson correlation between different metrics based on their evaluations of reports from CT-CHAT. As expected, lexical metrics (BLEU-2, ROUGE-L, BERTScore) show strong correlations with each other, as they all measure variations of surface or semantic similarity. Entity-based metrics (RadGraph, RaTEScore) also form a distinct cluster. Notably, CT-FineBench exhibits only a moderate correlation with these other metric families. This indicates that our benchmark is capturing a distinct and complementary signal related to fine-grained clinical accuracy, which is not fully represented by other evaluation paradigms.

\subsection{Ablation on Evaluation Model}

Our primary evaluation pipeline uses the powerful Qwen3-Max~\cite{yang2025qwen3} model for the QA and Answer Comparison steps in Section 3.3. To assess the feasibility of a more lightweight setup, we ablate this component and replace it with smaller, open-source models (Qwen3-32b and Qwen3-8b). The results are presented in Table~\ref{model_ablation}. The scores from the smaller models demonstrate a very high correlation with the scores from Qwen3-Max (e.g., Pearson's $\tau > 0.9$ for the 32b model). Furthermore, the absolute accuracy scores are remarkably close, with the 32b model achieving nearly identical performance to the Max model. This finding is highly encouraging, as it demonstrates that CT-FineBench can be deployed effectively using smaller, locally-runnable models without a significant loss in evaluation quality, making our benchmark both robust and practical for the wider research community.

\begin{table}[ht]
\centering
\resizebox{\columnwidth}{!}{
\begin{tabular}{lcccc}
\toprule
 & \multicolumn{4}{c}{\textbf{CT-RATE}} \\
\cmidrule(lr){2-5} 
 & \textbf{Pearson $\tau$} & \textbf{Kendall $\tau$} & \textbf{Spearman $\tau$} & \textbf{Acc} \\
\midrule
Qwen3-Max   & - & - & - &  15.8\\
Qwen3-32b  & 0.911 & 0.921 & 0.805 & 15.9 \\
Qwen3-8b   & 0.863 & 0.896 & 0.770 & 15.0 \\
\midrule
 & \multicolumn{4}{c}{\textbf{Merlin}} \\
\cmidrule(lr){2-5} 
 & \textbf{Pearson $\tau$} & \textbf{Kendall $\tau$} & \textbf{Spearman $\tau$} & \textbf{Acc} \\
\midrule
Qwen3-Max   & - & - & - &  22.4\\
Qwen3-32b  & 0.978 & 0.976 & 0.936 & 22.9 \\
Qwen3-8b   & 0.953 & 0.931 & 0.875 & 24.6 \\
\bottomrule
\end{tabular}}
\caption{Ablation on evaluation model} 
\label{model_ablation} 
\end{table}

\section{Conclusion}

We introduced CT-FineBench, a novel benchmark to address the failure of existing metrics in evaluating the fine-grained factual accuracy of generated CT reports. Our approach reframes evaluation as a Question-Answering task, verifying specific clinical attributes rather than relying on lexical or coarse entity matching. Experiments show that CT-FineBench is significantly more sensitive to fine-grained clinically errors and aligns more closely with human expert judgments than traditional metrics. The benchmark reveals that even state-of-the-art models struggle with correct fine-grained clinical details, highlighting a critical gap for clinical deployment. By providing a robust, interpretable evaluation standard, CT-FineBench paves the way for developing more clinically trustworthy and factually reliable report generation systems.

\section*{Limitations}
Our work has three primary limitations. First, since the question-answer pairs for CT-FineBench are constructed exclusively from the details present in the reference report, our evaluation is inherently recall-oriented. It excels at identifying errors of omission but does not penalize hallucinations or fabrications not related to the ground-truth findings. Therefore, CT-FineBench should be used in conjunction with other evaluation metrics, such as those that can measure precision, to provide a more comprehensive assessment. Second, although we construct CT-FineData, the training set version of CT-FineBench, we do not further explore its potential on improving model's fine-grained clinical accuracy. Third, the scope of our current benchmark is constrained by the predefined finding labels provided with the source datasets. We have not yet expanded our attribute schema to encompass all possible findings that may appear in the reports, which limits its coverage for unannotated findings.


\bibliography{custom}

\appendix

\section{Prompts and Guideline in Benchmark Construction}

This appendix provides the complete prompts and guidelines utilized during the benchmark construction phase, as detailed in Section 3.2. Our goal is to offer the reproducibility of the CT-FineBench creation process. We present three key components: 1) The prompt for Named Entity Recognition (NER), designed to automatically extract initial finding-attribute triplets from reference reports. 2) The detailed guideline for human annotators, which structured the critical task of refining the attribute schema to ensure clinical relevance and consistency. 3) The prompt for Question-Answer Pair Construction, used to systematically convert the structured schema and report content into the final benchmark data.

\section{Prompts in Experiment}

This appendix presents the specific prompts used during the experimental evaluation, as described in Section 3.3 and Section 5.2. These prompts are central to both our standard evaluation protocol and our sensitivity analysis. The section is organized as follows: 1) The Question Answering prompt, which instructs the model to extract answers from a candidate report. 2) The Answer Scoring prompt, which provides the LLM-based evaluator with the detailed criteria for assigning a score of 0, 0.5, or 1. 3) The prompts for generating Adversarial and Paraphrased Reports, which were used to create the test sets for the sensitivity analysis described in Section 5.2, designed to test a metric's ability to detect fine-grained errors while being robust to lexical variation.

\section{Evaluation Cost on Open-Source Models}

This section details the computational efficiency of our evaluation framework. All benchmarks were conducted on a single NVIDIA A800 GPU. To enhance performance, the inference process is accelerated using the VLLM library. The results demonstrate that our framework achieves a practical and acceptable time cost, making it suitable for large-scale evaluations. Table \ref{tab:throughput} summarizes the evaluation speed, measured in reports per second, for different models and benchmarks.

\begin{table}[h!]
\centering
\begin{tabular}{lcc}
\toprule
\textbf{Benchmark} & \textbf{Qwen3-8b} & \textbf{Qwen3-32b} \\
\midrule
CT-RATE   & 0.42    & 0.11     \\
Merlin    & 1.85    & 0.48     \\
\bottomrule
\end{tabular}
\caption{Evaluation time cost of our framework on a single A800 GPU. The values are measured in reports per second.}
\label{tab:throughput}
\end{table}

\begin{figure*}[t]
\begin{tcolorbox}[
    colback=black!5,  
    colframe=black!75, 
    title=Prompt for Named Entity Recognition, 
    fonttitle=\bfseries,
    arc=2mm, 
]
\noindent 
\textbf{CT Report:} \texttt{<report>} \\

\noindent
You are a Named Entity Recognition assistant. I will provide you with a CT imaging report, and you are to process it according to the following steps. No summaries or explanations: \\

\textbf{Step 1:} Perform fine-grained segmentation of the report as per the requirements.

1. Split each sentence of the report as finely as possible, ensuring that each resulting sentence describes only a single finding or disease. If an original sentence involves multiple findings or diseases, convert it into multiple sub-clauses, each corresponding to a different finding or disease.

2. Ensure that each segmented sentence is specific, clear, and does not use pronouns. If the original sentence contains pronouns, perform anaphora resolution during segmentation to clarify the subject being described, ensuring that details such as location, size, etc., are not omitted.

3. Print out each segmented sentence. Use <step1> and </step1> as tags to enclose the output. \\

\textbf{Step 2:} Classify the sentences segmented in Step 1 under each disease/finding from a given list, adhering to the following requirements:

1. A single key sentence may correspond to multiple diseases or findings.

2. If a specific disease or finding is mentioned, it may be described by multiple sentences. Record all corresponding sentences in a list.

3. If a mentioned disease or finding has no corresponding key description, output an empty list.

4. The output must be in JSON format. Use <step2> and </step2> as tags to enclose the output. \\

\textbf{Step 3:} Convert each short sentence from the JSON in Step 2 into one or more quadruplets to describe the details of the disease/finding. The format of a quadruplet is: (short sentence, finding/disease, attribute, value). The attribute can include size, location, shape, density, boundary, enhancement, etc. For example, the phrase "fibrotic lesions in both lower lungs" would be converted into a list like ["fibrotic lesions in both lower lungs", "fibrotic lesion", "location", "both lower lungs"]. The output should be a JSON-formatted list of these quadruplet lists. Use <step3> and </step3> as tags to enclose the output.

\end{tcolorbox}
\end{figure*}

\begin{figure*}[t]
\begin{tcolorbox}[
    colback=black!5,  
    colframe=black!75, 
    title=Guideline for Human Annotator in Attribute Definition, 
    fonttitle=\bfseries,
    arc=2mm, 
]
\noindent 

Your work is to identify a set of important attributes for each finding from CT Reports. You must ensure that every attribute has a clear medical meaning and is distinct from other attributes. 
Each file stores a series of raw attribute data for a single finding, which have been extracted through data mining and are sorted by attribute name. Each data entry for a attribute consists of four elements: finding, original report snippet, attribute, and attribute value. \\

You should use the specific content within this data (e.g., the report snippets and attribute values) to judge whether a mined attribute is medically significant, whether multiple different attributes can be merged into a single one, or whether a general attribute should be split into more granular ones. We recommend the following three-step annotation process:
(1) Prune medically insignificant attributes.
(2) Decompose general attributes into more specific attributes. Avoid single attributes that encompass multiple distinct characteristics.
(3) Merge similar attributes into one. For example, 'position' and 'location' can be consolidated. 
(4) Finally, after you have refined and finalized the attribute set, each attribute should be output on a new line, containing its attribute name, a description/explanation of the attribute, and a list of corresponding examples (please provide at least 5).

\end{tcolorbox}
\end{figure*}

\begin{figure*}[t]
\begin{tcolorbox}[
    colback=black!5,  
    colframe=black!75, 
    title=Prompt for Question-Answer Pair Construction, 
    fonttitle=\bfseries,
    arc=2mm, 
]
\noindent 
\textbf{CT Report:} \texttt{<report>} \\
\textbf{Disease/Finding:} \texttt{<disease>} \\
\textbf{Attribute:} \texttt{<attribute>} \\
\textbf{Attribute Explanation:} \texttt{<attribute explanation>} \\
\textbf{Attribute Example:} \texttt{<attribute example>} \\

\noindent
Given a CT report that contains a specific disease/finding. We also provide a key attribute of this disease/finding, an explanation of the attribute, and representative examples corresponding to this key attribute. Please follow the three steps provided below to generate the output, with output of each step on a new line. No summary or explanation is needed. \\

\textbf{Step 1:} Based on the CT report, determine if there is a description of the <attribute> for the <disease>. Follow these requirements:

1. If there is a corresponding description, output <step1>Yes</step1>.

2. If there is no corresponding description, output <step1>No</step1>. \\

\textbf{Step 2:} If the CT report describes the <attribute> for the <disease>, refer to the attribute explanation and attribute examples, and extract the corresponding attribute content. Follow these requirements:

1. If there is no corresponding description as determined in Step 1, output <step2>[]</step2>.

2. If there is a corresponding description as determined in Step 1, output a pair: <step2>[attribute, extracted attribute content]</step2>, for example, <step2>["location", "left lung"]</step2>.

3. Multiple different attribute examples are provided here, separated by commas. Note that the attribute examples do not represent all possible options for the attribute content; they are only representative examples.

4. The extracted attribute content should be concise and accurate. \\

\textbf{Step 3:} Based on the disease, the attribute, and the extracted content from Step 2, transform them into a question-answer pair. The purpose of the question is to inquire about the status of this attribute in the report, and the answer is the extracted attribute content. Follow these requirements:

1. If there is no corresponding description as determined in Step 1, output <step3>[]</step3>.

2. If there is a corresponding description as determined in Step 1, output a pair: <step3>[question, answer]</step3>, for example, <step3>["Where is the location of lung opacity?", "left lung"]</step3>.

3. The question must contain information about the disease and the attribute being queried, clearly pointing to the key attribute, but must not contain the answer content or its synonyms. The question should use general terminology and avoid directly quoting specific words from the report.

4. The answer must be exactly the same as the extracted attribute content from Step 2.

\end{tcolorbox}
\end{figure*}

\begin{figure*}[t]
\begin{tcolorbox}[
    colback=black!5,  
    colframe=black!75, 
    title=Prompt for Question Answering, 
    fonttitle=\bfseries,
    arc=2mm, 
]
\noindent 
\textbf{CT Report:} \texttt{<report>} \\
\textbf{Question:} \texttt{<question>} \\
\textbf{Explanation:} \texttt{<explanation>} \\
\textbf{Example:} \texttt{<example>} \\[1em]

\noindent
Given a CT report, you need to answer the given question based on the CT report. The question is about the \texttt{<attribute>} of \texttt{<disease>}. We also provide some information about \texttt{<attribute>}, including the explanation of \texttt{<attribute>}, and some representative examples of \texttt{<attribute>}. Please answer the input question based on CT report and the other information. If the CT report do not contain any information about the question, just return 'no answer'. \\[1em]
\end{tcolorbox}
\end{figure*}

\begin{figure*}[t]
\begin{tcolorbox}[
    colback=black!5,      
    colframe=black!75,     
    title=Prompt for Answer Scoring, 
    fonttitle=\bfseries,   
    arc=2mm,               
    boxrule=0.5pt,         
]

\medskip 
\noindent\textbf{Score 1 (Correct / Fully Acceptable):}
The predict answer is fully correct and semantically equivalent to the reference answer, or provides a correct, more detailed version of it.

\textbf{Exact or Semantic Match:} The answer is identical or uses different wording (synonyms, rephrasing) to convey the exact same meaning.

\textbf{Correct but More Specific:} The predict answer is a correct, but more specific, instance of the reference answer.

\textbf{Contains Correct Additional Detail:} The predict answer includes all key information from the reference and adds other correct, relevant details.

\textbf{Numerical Answers:} When both answers are numbers, the predict answer matches exactly or is within a very close tolerance (e.g., ±10\%) of the reference answer.

\noindent\textbf{Score 0.5 (Partially Correct):}
The predict answer is on the right track but is flawed by being incomplete or too general.

\textbf{Correct but Incomplete:} The predict answer provides correct information but omits some key elements from the reference answer.

\textbf{Correct but Overly General:} The predict answer is a correct but less specific version of the reference answer. It captures the essence but loses important detail.

\textbf{Numerical Answers:} The predict answer is numerically in the same ballpark (same order of magnitude) but outside the strict tolerance for a score of 1(10\% < difference < 30\%).

\noindent\textbf{Score 0 (Incorrect):}
The predict answer is factually wrong, irrelevant, or fails to answer the question.

\textbf{Contradiction:} It directly contradicts the reference answer.

\textbf{Wrong Information:} It provides a completely different piece of information.

\textbf{Hallucination:} It provides information not supported by the context or states that an answer cannot be determined when the reference provides one.

\textbf{Numerical Answers:} The predict answer is of a different order of magnitude or significantly incorrect (difference > 30\%).

\subsection*{Instruction}
You are an expert evaluator for medical question answering systems. Your task is to score a predict answer against a reference answer for a given question.
\par
You must evaluate the semantic accuracy and completeness of the predict answer with the above scoring criteria. Your final output must consist of a score from the set \{0, 0.5, 1\}.

\end{tcolorbox}
\end{figure*}

\begin{figure*}[t]
\begin{tcolorbox}[
    colback=black!5,      
    colframe=black!75,     
    title=Prompt for Adversarial Report, 
    fonttitle=\bfseries,   
    arc=2mm,               
    boxrule=0.5pt,         
]

\subsection*{Background}
Your task is to act as an expert clinical data scientist creating a comprehensive suite of adversarial examples from a single medical report. Your goal is to autonomously identify all potential points of critical clinical failure within a given reference report and, generate a distinct modified version of the report with clinically error.

\subsection*{Instruction}
The input is a CT report. You need to completely change details of the report, while maintaining the overall similarity. Details include the position, size, margin, shape, severity, appearance, morphology, density, associated signs, etiology, type, distribution, pattern and etc. However, the details do not include existence of any finding/disease or whether a organ is abnormal. For example, you can not change "Thoracic esophagus calibration was normal" to "Thoracic esophagus calibration was abnormal". The change for details is not for improving it, but convert it to an opposite direction. Change as much details as you can, but still with high n-gram similarity. Do not change the format of the input report.

\end{tcolorbox}
\end{figure*}

\begin{figure*}[t]
\begin{tcolorbox}[
    colback=black!5,      
    colframe=black!75,     
    title=Prompt for Paraphrased Report, 
    fonttitle=\bfseries,   
    arc=2mm,               
    boxrule=0.5pt,         
]

\subsection*{Background}
Your task is to act as an expert medical professional specializing in clinical report writing and linguistic analysis. Your goal is to create a semantically equivalent but stylistically divergent version of a given medical report. This process is designed to rigorously test the robustness of medical report evaluation metrics against linguistic and stylistic variations. The generated "Modified Report" should be clinically identical to the original but use different vocabulary, phrasing, and sentence structure to be textually as dissimilar as possible. 

\subsection*{Instruction}
The input is a CT report. You need to extensively paraphrase the report to minimize n-gram similarity with the original, while ensuring the clinical meaning remains absolutely unchanged. You must **not** alter any clinical details or diagnostic conclusions. Details that must be preserved include: the existence/absence of any finding/disease, diagnosis, measurements (size), location (position), characteristics (e.g., margin, shape, density, morphology, appearance), severity, and any described etiology, type, distribution, or pattern. The evidence leading to the diagnosis must be identical. For example, you cannot change "Thoracic esophagus calibration was normal" to "Thoracic esophagus calibration was abnormal". Your main objective is to use synonyms, reorder clauses, and change sentence structures to make the text as different as possible from the original, thereby achieving low n-gram similarity. Do not change the format of the input report

\end{tcolorbox}
\end{figure*}
\end{document}